\documentclass[12pt]{article}
\usepackage{amsmath}
\usepackage{times}
\usepackage{graphicx}
\usepackage{color}
\usepackage{multirow}
\usepackage[authoryear]{natbib}
\usepackage{rotating}
\usepackage{bbm}
\usepackage{latexsym}

\usepackage{subfigure} 

\usepackage{algorithm}
\usepackage{algorithmic}

\usepackage{amssymb}
\usepackage{amsmath}
\usepackage{graphicx}
\usepackage{color}
\usepackage{caption}
\usepackage{natbib}
\usepackage{url}

\newcommand{\disT}{\textstyle}
\newcommand{\disS}{\displaystyle}
\newcommand{\prob}[2]{p(#1 \, | \, #2)}  
\newcommand{\Prime}{\,'}  
\newcommand{\One}{I}
\renewcommand{\vec}[1]{{\mathbf{#1}}}
\newcommand{\Kn}{\mathcal{K}_{n}}

\newcommand{\Ss}{\mathcal{S}}

\usepackage{float}

\newcommand{\captionfonts}{\normalsize}

\makeatletter  
\long\def\@makecaption#1#2{%
  \vskip\abovecaptionskip
  \sbox\@tempboxa{{\captionfonts #1: #2}}%
  \ifdim \wd\@tempboxa >\hsize
    {\captionfonts #1: #2\par}
  \else
    \hbox to\hsize{\hfil\box\@tempboxa\hfil}%
  \fi
  \vskip\belowcaptionskip}
\makeatother   

\begin{document}
\hspace{13.9cm}1

\ \vspace{20mm}\\ 

{\LARGE GP-select: Accelerating EM using adaptive \\subspace preselection}

\ \\
{\bf \large Jacquelyn A. Shelton$^{\displaystyle 1}$, Jan Gasthaus$^{\displaystyle 2,3}$, Zhenwen Dai$^{\displaystyle 4}$, J\"org L\"{u}cke$^{\displaystyle 5}$, and Arthur Gretton$^{\displaystyle 2}$}\\
{$^{\displaystyle 1 \,}$Technical University Berlin, Marchstrasse 23, 10587 Berlin, Germany.}\\
{$^{\displaystyle 2 \,}$University College London, Gatsby Unit, 25 Howland Street, London W1T 4JG, UK.}\\
{$^{\displaystyle 3 \,}$Amazon Development Center, Karl-Liebknecht-Str. 5, 10178 Berlin, Germany.}\\
{$^{\displaystyle 4 \,}$University of Sheffield, Western Bank, Sheffield, South Yorkshire S10 2TN, UK.}\\
{$^{\displaystyle 5 \,}$University of Oldenburg, Ammerl\"ander Heerstr. 114, 26129 Oldenburg, Germany.}\\
%

\thispagestyle{empty}
\markboth{}{NC instructions}
\ \vspace{-0mm}\\
%
\begin{center} 
{\bf Abstract} 
\end{center}
We propose a nonparametric procedure to achieve fast inference in generative graphical models when the number of latent states is very large.
 The approach is based on iterative latent variable preselection, where we alternate between learning a 'selection function' to reveal the relevant latent variables, and using this to obtain a compact approximation of the posterior distribution for EM; this can make inference possible where the number of possible latent states is e.g. exponential in the number of latent variables, whereas an exact approach would be computationally infeasible.
We learn the selection function entirely from the observed data and current EM state via Gaussian process regression. This is by contrast with earlier approaches, where selection functions were manually-designed for each problem setting.
We show that our approach performs as well as these bespoke selection functions on a wide variety of inference problems: in particular, for the challenging case of a hierarchical model for object localization with occlusion, we achieve results that match a customized state-of-the-art selection method,  at a far lower computational cost.




\definecolor{Green}{rgb}{0,.80,0} 

\section{Introduction}
%

Inference in probabilistic graphical models can be challenging in situations where
there are a large number of hidden variables, each of which may take on one of several
state values. The Expectation Maximization (EM) algorithm is widely applied for inference when hidden variables
are present, however inference can quickly become intractable as the dimensionality of hidden states increases.

Expectation truncation (ET) \citep{LuckeEggert2010} is a meta algorithm for accelerating EM learning
in graphical models, which restricts the inference performed during the E-step
to an ``interesting'' subset of states of the latent variables,  
chosen per data point according to a \emph{selection function}.
This subspace reduction can lead to a significant decrease in computation with very little loss of accuracy
(compared with the full model).
In previous work, functions to select states of high posterior mass were 
derived individually for each model of interest, e.g. by taking upper bounds or noiseless limits. 
\citep{LuckeEggert2010,SheltonEtAl2012,BornscheinEtAl2013,SheikhEtAl2014}.
The crucial underlying assumption remains that 
when EM has converged,
the posterior mass is  concentrated in small volumes of the latent state space.
This property is observed to hold in many visual data, auditory data, and general pattern recognition settings.

For more complex graphical models, notably the hierarchical model of \citet{DaiLucke2014},  the design of suitable selection functions
is extremely challenging: it requires both expert knowledge
on the problem domain and considerable computational resources to implement
 (indeed, the design of such functions for  particular problems
has been a major contribution in previous work on the topic).

In the present work, we propose a generalization of the ET approach, where
we avoid completely the challenge of problem-specific selection function design.
Instead,
we learn selection functions  adaptively and nonparametrically
from the data,
 while learning the model
parameters  simultaneously using EM.
We emphasize that the selection function is  used only to ``guide'' the underlying
base inference algorithm to regions of high posterior probability, but is not itself
used as an approximation to the posterior distribution. As such, the learned
function does not have to be a completely accurate indication of latent
variable predictivity,
as long as the relative importance of states is preserved.
We use  Gaussian process
regression \citep{RasmussenGPbook} to learn the selection function, though other regression techniques could
also be applied. The main advantage of GPs
is that they have an analytic one-shot learning procedure, and that learning different
functions based on the same inputs is computationally cheap, which makes adaptive learning
of a changing target function efficient. We term this part of our approach
\textit{GP-select}.
Our nonparametric generalization of  ET may be applied as a black-box
meta algorithm for accelerating inference in  generative graphical models,
with no expert knowledge required.





Our approach is the first to make ET a general purpose algorithm for discrete latents,
 whereas previously new versions of ET were required for every new latent variable model addressed. 
For instance, in Section \ref{invec} we will show that preselection is crucial for efficient inference in complex models. 
Although ET has already been successful in some models, this work shows that more complex models will crucially depend on an improved selection step and focuses on automating this step.

For empirical evaluation, 
we have applied GP-select in a number of experimental settings.
First, we considered the case of sparse coding models (binary sparse coding,
spike-and-slab, nonlinear spike-and-slab), where the relationship between the
observed and latent variables is known to be complex and nonlinear.\footnote{Note that
 even when linear relations exist between the latents and outputs, a nonlinear
regression may still be necessary in finding relevant variables,
as a result of explaining away.}
We show that GP-select can produce results with equal performance to the respective manually-derived selection functions.
Interestingly, we find it can be essential to use nonlinear GP regression
in the spike-and-slab case, and that simple linear regression is not
sufficiently flexible in modeling the posterior shape.
Second, we illustrate GP-select on a simple Gaussian mixture model,
where we can explicitly visualize the form of the learned regression function.
We find that even for a simple model, it can be be essential to learn a nonlinear mapping.
%
Finally, we present results
for a recently published hierarchical model for  translation invariant occlusive components analysis
\citep{DaiLucke2014}.
The performance of our inference algorithm matches that of the complex
hand-engineered selection function of the previous work, while being straightforward
to implement and having a far lower computational cost.




\section{Related work}


The general idea of aiding inference in graphical models by
learning a function that maps from the observed data to
a property of the latent variables is quite old. Early work includes the
Helmholtz machine \citep{Dayan95} and its bottom-up connections trained using the wake-sleep
algorithm \citep{HintonEtAl1995}.
More recently, the idea has surfaced in the context of learning variational distributions with neural networks~\citep{WellingICML2014}.
A two-stage inference has  been discussed in the context of
computer vision \citep{YuilleKersten2006} and neural inference \citep{KoernerEtAl1999}.
Recently, researchers~\citep{MnihGregor2014} 
have generalized this idea to learning in arbitrary graphical models by training
an ``inference network'' that efficiently implements sampling from the posterior
distribution.


GPs have recently been widely used to "learn" the results of complicated models in order to accelerate inference and parameter selection. 
GP approximations have been used in lieu of solving complex partial differential equations \citep{SacksEtal1989, CurrinEtall1991}, to learn data-driven kernel functions for recommendation systems \citep{SchwaighoferEtAl2005}, and recently for quantum chemistry \citep{RuppEtAl2012}. 
%
Other work has used GPs to simplify computations in approximate Bayesian computation (ABC) methods: namely to model the likelihood function for inference \citep{Wilkinsons2014}, to aid in making Metropolis-Hastings (MH) decisions \citep{MeedsWelling2014}, and to model the discrepancies between simulated/observed data parameter space simplification \citep{GuttmanCorander2015}.
Recently, instead of the typical choice of GPs for large scale Bayesian optimization, neural networks have been used to learn an adaptive set of basis functions for Bayesian linear regression \citep{SnoekEtAl2015}.

Our work follows the same high level philosophy in that we use GPs to approximate complex/intractable probabilistic models. None of the cited prior work address our problem setting, namely the selection of relevant latent variables by learning a nonparametric relevance function, for use in expectation truncation (ET).

\section{Variable selection for accelerated inference}
\label{method}
%
\textbf{Notation.}
We denote the observed data by the $D\times N$ matrix $\vec{Y}=(\vec{y}^{(1)}, \dots, \vec{y}^{(N)})$, where each vector $\vec{y}^{(n)} = ( y_1^{(n)}, \dots, y_D^{(n)})^\mathrm{T}$ is the $n$th observation 
in a $D$-dimensional space.
Similarly we define corresponding 
binary latent variables 
by the matrix $\vec{S} = (\vec{s}^{(1)}, \dots, \vec{s}^{(N)})\in \{0,1\}^{H \times N}$ 
where each $\vec{s}^{(n)}=(s_1^{(n)}\dots, s^{(n)}_H)^\mathrm{T} \in \{0,1\}^{H}$ is the $n$th vector in the $H$-dimensional latent space,
and for each individual hidden variable $h=1,\dots,H$, the vector $\vec{s}_h=(s_h^{(1)}\dots, s^{(N)}_h)\in \{0,1\}^{N}$. 
Reduced latent spaces are denoted by $H'$, where $H' \ll H$. 
Note that although we restrict ourselves to binary latent variables here, 
the procedure could in principle be generalized to variables with higher cardinality \citep[e.g. see]{ExarchakisEtAl2012}.
We denote the prior distribution over the latent variables with $p(\vec{s} | \theta)$ 
and the likelihood of the data with $p(\vec{y} | \vec{s}, \theta)$.
Using these expressions, the posterior distribution over latent variables is 
\vspace{-.1cm}
\begin{equation}
\label{eq:post}
p(\vec{s}^{(n)}|\vec{y}^{(n)},\Theta)  = \frac{p(\vec{s} | \Theta) \, p(\vec{y}^{(n)} | \vec{s}^{(n)}, \Theta)}
{\disS\hspace{-1.5mm}\sum_{\vec{s}\Prime^{(n)}} p(\vec{s}\Prime | \Theta) \, p(\vec{y} | \vec{s}\Prime, \Theta)}.
\end{equation}
\vspace{-.5cm}

\subsection{Selection via Expectation Truncation in EM}
Expectation Maximization (EM) is an iterative algorithm to optimize the model parameters of a given graphical model (see e.g. \citep{DempsterEtAl1977, NealHinton1998}).
EM iteratively optimizes a lower bound on the data likelihood by inferring the
posterior distribution over hidden variables given the current parameters (the
E-step), and then adjusting the parameters to maximize the likelihood of the
data averaged over this posterior (the M-step).
When the number of latent states to consider is large (e.g.\ exponential in the
number of latent variables), the computation of the posterior distribution in
the E-step becomes intractable and approximations are required.


%
Expectation truncation (ET) is a meta algorithm, which improves convergence of the expectation maximization (EM) algorithm \citep{LuckeEggert2010}.
The main idea underlying ET is that the posterior probability mass is concentrated in a small subspace of the full latent space.
This is the case, for instance, if for a given data point $\vec{y}^{(n)}$ 
only a subset of the $H$ latent variables $s_h^{(n)}$ are relevant. Note, however, that the posterior may still be concentrated  even when all latents are relevant, since most of the probability mass may be concentrated on few of these. 

A \textit{selection function} can be used to identify a  subset
of salient variables, denoted by $H'$ where $H' \ll H$, which in turn is used to define a subset, denoted $\mathcal{K}_n$, of the possible state configurations of the space per data point. 
State configurations not in this space (of variables deemed to be non-relevant) are fixed to $0$ (assigned zero probability mass).
The posterior distribution~\eqref{eq:post} can then be approximated by a truncated posterior distribution, computed on the reduced support,
\vspace{-.1cm}
\begin{align}
\label{eq:sel-post}
p(&\vec{s}^{(n)}|\vec{y}^{(n)},\Theta) \nonumber\\
&\approx q_n(\vec{s}^{(n)};\Theta) = \frac{p(\vec{s}^{(n)},\vec{y}^{(n)}|\,\Theta) \,\delta(\vec{s}^{(n)}\in\,\,\Kn)}
{\disS\hspace{-1.5mm}\sum_{\vec{s}\Prime^{(n)}\in\mathcal{K}_n}\hspace{-3mm} p(\vec{s}\Prime^{(n)},\vec{y}^{(n)}|\,\Theta)},
\end{align}
\normalsize
where $\Kn$ contains the latent states of the $H'$ relevant variables for data point
$\vec{y}^{(n)}$, and $\delta(\vec{s}\in\mathcal{K}_n)=1$ if
$\vec{s}\in\mathcal{K}_n$ and zero otherwise.
In order words, Eq.~\eqref{eq:sel-post} is proportional to Eq.~\eqref{eq:post} if $s \in \Kn$ (and zero otherwise). 
The set $\Kn$ contains only states for which $s_h=0$ for all $h$ that are not selected, i.e. all states where $s_h=1$ for non-selected $h$ are assigned zero probability.
The sum over $\Kn$ is much more efficient than the sum for the full posterior, since it need only be computed over the reduced set of latent variable states deemed relevant: the state configurations of the irrelevant variables are fixed to be zero.
The variable selection parameter H' is selected based on compute resources available: i.e. as large as resources allow in order to be closer to true EM, although empirically it's been shown that much smaller values suffice  \citep[see e.g.][App. B on complexity-accuracy trade-offs]{SheikhEtAl2014}.


%

\subsection{ET with affinity} 

\begin{figure}[h]
\begin{center}
\includegraphics[width=.535\textwidth]{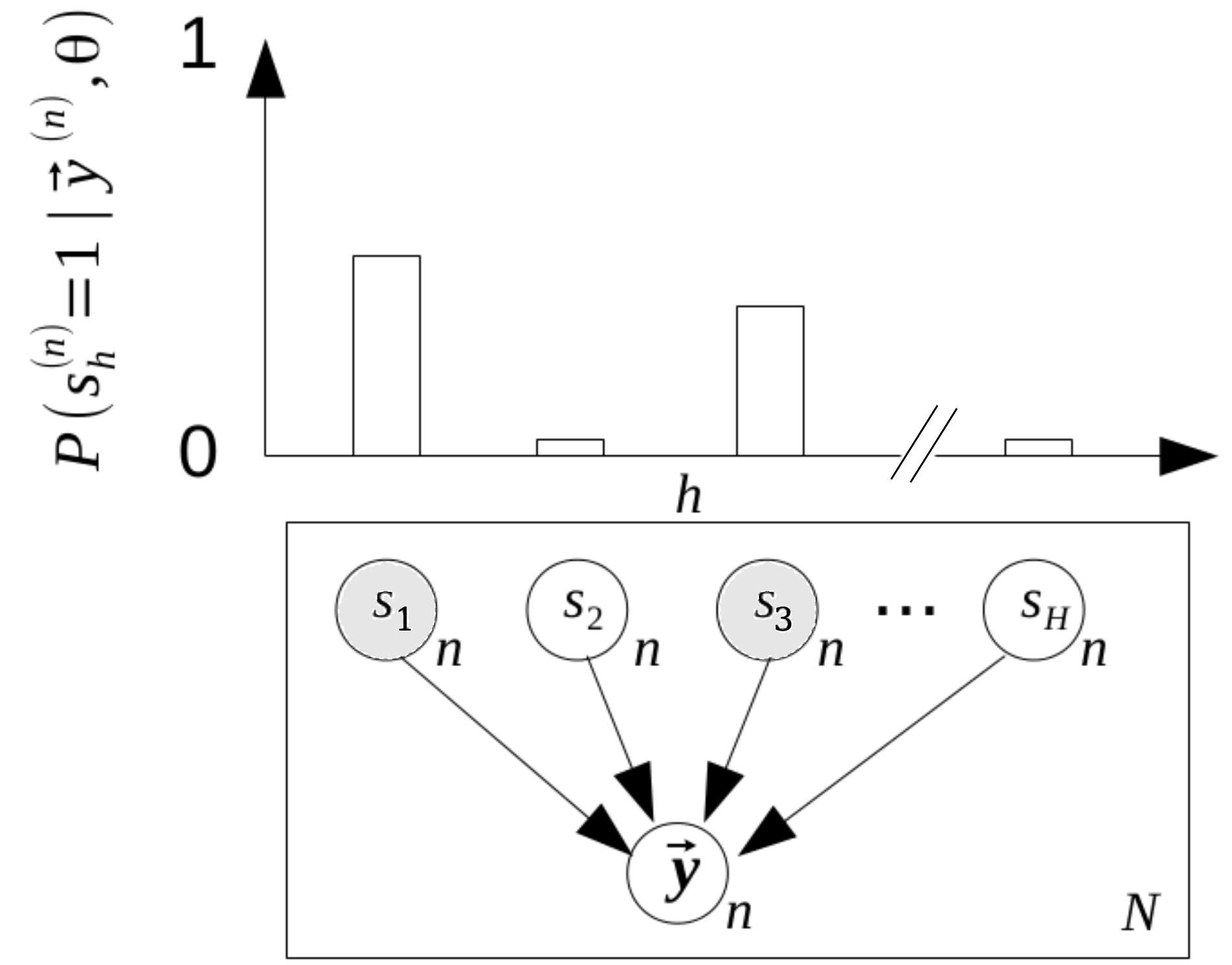}
\caption{Illustration of the affinity function for selection.
The affinity approximates the marginal posterior probability of each $h=1,\dots,H$ latent variable (top), 
which corresponds to the most relevant variables for a given data point $\vec{y}^{(n)}$ (bottom). 
Here, the affinity would return variables $s_1$ and $s_3$ as being relevant for $\vec{y}^{(n)}$. 
}\label{fig:graph-affinity}
\end{center}
\end{figure}
%
One way of constructing a selection function 
is by first ranking the latent variables according to an 
\emph{affinity function} $f_h(\vec{y}^{(n)}) : \mathrm{R}^D \mapsto \mathrm{R}$ 
which directly reflects the relevance of latent variable $s_h$. 
A natural choice for such a function is the one that approximates the marginal posterior probability 
of each variable, e.g.\ we try to learn $f$ as follows:
\begin{equation}
\label{eq:affinity}
f_h(\vec{y}^{(n)}) = \hat{p}_h^{(n)} \approx p^{(n)}_h \equiv p(s^{(n)}_h = 1|\vec{y}^{(n)}, \Theta),
\end{equation}
meaning that, for the relevant variables, the marginal posterior probability $p_h$ 
exceeds some threshold. 
See Figure~\ref{fig:graph-affinity} for a simplified illustration. 
When the latent variables $s_{h=1}^{(n)}, \dots, s_H^{(n)}$ in the marginal posterior probability $\hat{\vec{p}}^{(n)} = \hat{p}_{h=1}^{(n)},\dots, \hat{p}_H^{(n)}$ are conditionally independent given a data point $\vec{y}^{(n)}$, this affinity function correctly isolates the most relevant variables in the posterior.
Even when this strong assumption does not hold in practice (which is often the case), however,
the affinity can still correctly highlight relevant variables,
and has been empirically shown to be quite effective even when dependencies exist (see e.g. source separation tasks in \citep{SheikhEtAl2014}).

%
Next, using all $\hat{p}_{h=1}^{(n)},\dots, \hat{p}_H^{(n)}$  
from the affinity function 
$\vec{f}(\vec{y}^{(n)}) = (f_1(\vec{y}^{(n)}), \dots, f_H(\vec{y}^{(n)}))$, we define 
 $\gamma\,(\hat{\vec{p}}^{(n)})$ to simultaneously sort the indices of the latent variables in descending order and \textit{reduce} the sorted set to the $H'$ highest (most relevant) variables' indices. 
To ensure that there is a non-zero probability of selecting each variable per EM iteration, $10\%$ of the $H'$ indices are uniformly chosen from $H$ at random. 
This prevents the possible propagation of errors from $q_{(n)}$ continuously assigning small probabilities to a variable $s_h$ in early EM
iterations. 
$\gamma(\hat{\vec{p}}^{(n)})$ thus returns the $H'$ selected variable indices $I$ deemed by the affinity to be relevant to the $n$th data point.
%
%
%
Finally, using the indices $I$ from $\gamma$, we define $\mathcal{I}(I)$ to return an 
$H'$-dimensional subset of selected relevant latent states $\Kn$ for each data point $\vec{y}^{(n)}$. 
All `non-relevant' variable states $s_h$ for all variables $h\not\in I$ are effectively set to $0$ in Equation~\eqref{eq:sel-post} 
by not being present in the state set $\Kn$.
%
%

Using $\vec{f}$, $\mathcal{I}$, and $\gamma$, we can define a 
\emph{selection function} $\mathcal{S}: \mathrm{R}^D \mapsto 2^{ \{1,\dots,H \}}$ 
to select subsets $\mathcal{K}_n$ per data point $\vec{y}^{(n)}$.   
Again, the goal is for the states $\Kn$ to contain most of the probability mass
$\prob{\vec{s}}{\vec{y}}$ and to be significantly smaller than the entire
latent space.  
The \textit{affinity based selection function}  can be expressed as
\vspace{-.1cm}
\begin{equation}\label{eq:sel-func}
\mathcal{S}(\vec{y}^{(n)}) \;=\; \mathcal{I} \left[  \gamma \left[ \vec{f}(\vec{y}^{(n)}) \right]  \right] \;=\; \mathcal{K}_n.
\end{equation}
%


\subsection{Inference in EM with selection}
%
In each iteration of EM, the following occurs: 
prior to the E-step, the selection function $\Ss(\vec{y}^{(n)})$ in \eqref{eq:sel-func} is computed to select the most relevant states $\Kn$, 
which are then used to compute the truncated posterior distribution $q_n(\vec{s})$ in \eqref{eq:sel-post}.
The truncated posterior can be computed using any standard inference method, 
such as exact inference or e.g. Gibbs sampling from $q(\vec{s})$  
if inference is still intractable or further computational reduction is desired.
The result of the E-step is then used to update the model parameters in the M-step. 


\section{GP-Select}
\label{gp-select}
%
In previous work, the selection function $\mathcal{S}(\vec{y}^{(n)})$ 
was a deterministic function derived  individually for each model 
\citep[see e.g.][]{SheltonEtAl2011, SheltonEtAl2012, DaiLucke2012a, DaiLucke2012b,
BornscheinEtAl2013, SheikhEtAl2014, SheltonEtAl2015}.
We now generalize the selection approach:  
instead of predefining the form of $\Ss$ for variable selection, we want
to learn it in a black-box and model-free way based on the data.
We learn $\Ss$  using Gaussian process (GP) regression
\citep[e.g.][]{RasmussenGPbook}, which is a flexible nonparametric model 
and scales cubicly\footnote{If the scaling with $N$ is still too expensive, an incomplete Cholesky approximation is used, with cost linear in $N$ and quadratic in the rank $Q$ of the approximation (see Section~\ref{invec} for details).} with the number of data points $N$ but linearly with the number of latent variables $H$.  
%
%
%
%
We define the affinity function $f_h$ as being drawn from a Gaussian process model: 
$f_h(\vec{y}^{(n)}) \sim \text{GP}\left(0, \, k(\cdot,\cdot) \right)$, where $k(\cdot, \cdot)$ is the covariance kernel, 
which can be flexibly parameterized to represent the relationship between variables.
Again, we use $f_h$ to approximate the marginal posterior probability $p_h$ that $s_h^{(n)}=1$.
A nice property of Gaussian processes is that the kernel matrix $K$ need only be computed once (until the kernel function hyperparameters are updated) 
to approximate $p_h^{(n)}$ for the entire $H\times N$ set of latent variables $\vec{S}$.

Thus, prior to each E-step in each EM iteration, within each calculation of the selection function, we calculate the affinity  using a GP to regress the expected values of the latent variables $\langle \vec{S} \rangle$ onto the observed data $\vec{Y}$.  
Specifically, we train on $p_h$ from the previous EM iteration (where $p_h$ is equal to $\langle s_h \rangle$), for 
training data of 
$\mathcal{D} = \{ (\vec{y}^{(n)}, \langle\vec{s}^{(n)}\rangle_{q}) | n = 1,\dots, N \}$, 
where we recall that $q_{n}(\vec{s}^{(n)})$ is the approximate posterior distribution for $\vec{s}^{(n)}$ in Eq.~\eqref{eq:sel-post}.
%
In the first EM iteration, the expectations $\langle\vec{s}^{(n)}\rangle_{q}$ are initialized randomly;
in each subsequent EM iteration, the 
expectations w.r.t. the $\Kn$-truncated posterior $q(\vec{s})$ are used. 
The EM algorithm is run for $T$ iterations and the hyperparameters of the kernel are optimized by maximum likelihood every $T^*$ EM iterations.

For each data point $n$ and latent variable $h$ we compute the predicted mean of the GP by leaving 
this data point out of the training set and considering all others, which is called leave-one-out (LOO) prediction.
It can be shown that this can be implemented efficiently \citep[see Section 5.4.2 in ][]{RasmussenGPbook}, 
and we use this result to update the predicted affinity as follows:
%
%
\vspace{-.2cm}  
\begin{equation}\label{eq:gp-loo}
\hat{p}_{h}^{(n)} \leftarrow   
\langle s_h^{(n)}\rangle_{q_n} - \frac{ [ K^{-1} \langle\vec{s}_{h}\rangle_{q_n} ]_{nn} }{ [ K^{-1} ]_{nn} }.
\end{equation}
%
%
Equation~\eqref{eq:gp-loo} can be efficiently implemented for all latent variables $h=1,\dots,H$ and all data points $n=1,\dots,N$ using matrix operations, thereby requiring only one kernel matrix inversion for the entire dataset.
%
%
%

Substituting Equation~\eqref{eq:gp-loo} for $\vec{f}$ in the affinity based selection function~\eqref{eq:sel-func}, 
we call the entire process \textit{GP-select}. An outline is shown in Algorithm 1.

\begin{algorithm}
\caption{GP-Select to accelerate inference in Expectation Maximization}
\label{alg:gp-select}
\begin{algorithmic}
\FOR{EM iterations $t=1,\dots,T$}\STATE{
    \FOR{data point $n=1,\dots, N$}
    \STATE{
    compute affinity of all latent variables $\hat{\vec{p}}_t^{(n)}$: \eqref{eq:gp-loo}\\
    compute subset of relevant states $\Ss$: \eqref{eq:sel-func}\\ 
    compute truncated posterior $q_{n,t}(\vec{s}^{(n)})$, E-step: \eqref{eq:sel-post}\\ 
    update model parameters in M-step\\ 
    store $\langle s_h^{(n)}\rangle_{q_{n,t}}$ for $\vec{p}^{(n)}$ in EM iteration ${t+1}$\\
    }\ENDFOR\\
    optimize kernel hyperparams every $T^{*}$ EM iterations
}
\ENDFOR
\end{algorithmic}
\end{algorithm}

\section{Experiments}
\label{exps}
%
We apply our GP-select inference approach to five different probabilistic generative models.
First, we considered three sparse coding models (binary sparse coding,
spike-and-slab, nonlinear spike-and-slab), where the relationship between the
observed and latent variables is known to be complex and nonlinear.
Second, we apply GP-select to a simple Gaussian mixture model,
where we can explicitly visualize the form of the learned regression function.
%
Finally, we apply our approach to a recent hierarchical model for translation invariant occlusive components analysis
\citep{DaiLucke2012a,DaiEtAl2013,DaiLucke2014}.


\subsection{Sparse coding models}
\label{sec:sparse-coding}
Using hand-crafted functions to preselect latent variables, a variety of sparse coding models have been successfully scaled to high-dimensional  latent spaces with use of  selection~\citep{HennigesEtAl2010, BornscheinEtAl2013, SheikhEtAl2014} and selection with Gibbs sampling~\citep{SheltonEtAl2011, SheltonEtAl2012, SheltonEtAl2015} inference approaches.
In order to demonstrate our method, we consider three of these
 sparse generative models,  and perform inference with our GP-select approach instead of a hand-crafted selection function. The models are:
%
\begin{description}
\item[\textbf{A.}] \textit{Binary sparse coding}:
\vspace{-.1cm}
\begin{align}
\text{latents: } \,\,\,\vec{s} \sim Bern(\vec{s} | \pi) &= \disT\prod_{h=1}^H \pi^{s_h} \big( 1 - \pi \big)^{1-s_h},\nonumber \\
\text{observations:  }\,\,\,    \vec{y} &\sim \mathcal{N}(\vec{y}; W\vec{s}, \sigma^2\One),\nonumber
\end{align}
where $W \in \mathrm{R}^{D \times H}$ denotes the dictionary elements and $\pi$ parameterizes the sparsity (see e.g.~\citep{HennigesEtAl2010}).
\item[\textbf{B.}] \textit{Spike-and-slab sparse coding}:
\vspace{-.2cm}
\begin{align}
\text{latents: } & \,\,\,\vec{s} = \vec{b}\odot\vec{z}
\quad\mathrm{where}\quad \vec{b}\sim Bern(\vec{b} | \pi)
\:\mathrm{and} \:
\vec{z} \sim \mathcal{N}(\vec{z};\,\vec{\mu} , \Sigma_h),\nonumber\\
\text{observations:} & \,\,\, \vec{y} \sim \mathcal{N}(\vec{y}; W\vec{s}, \sigma^2\One)\nonumber
\end{align}
where the point-wise multiplication of the two latent vectors, i.e., $(\vec{s}\odot\vec{z})_h = s_h\,z_h$
generates a `spike-and-slab' distributed variable ($\vec{s}\odot\vec{z}$), that has either continuous values or exact zero entries (e.g. \citep{TitsiasGredilla2011,GoodfellowEtAl2013,SheikhEtAl2014}).
\item[\textbf{C.}] \textit{Nonlinear Spike-and-slab sparse coding}:
\vspace{-.2cm}
\begin{align}\label{eq:ssmca}
\text{latents: } & \,\,\,\vec{s} = \vec{b}\odot\vec{z}
\quad\mathrm{where}\quad \vec{b}\sim Bern(\vec{b} | \pi)
\:\mathrm{and} \:
\vec{z} \sim \mathcal{N}(\vec{z};\,\vec{\mu} , \Sigma_h),\nonumber\\
 \text{observations:}\,\,\, &\vec{y} \sim \mathcal{N}(\vec{y}; \max_{h}\{s_h\vec{W}_{h}\}, \sigma^2\One)\nonumber
\end{align}
for which the mean of the Gaussian for each $\vec{y}^{(n)}$ is centered at $\max_{h}\{s_h\vec{W}_{h}\}$, where $\max_{h}$ is a nonlinearity that considers all $H$ latent components and takes the $h$ yielding the maximum value for $s_h\vec{W}_{h}$ \citep{LuckeSahani2008,SheltonEtAl2012,BornscheinEtAl2013,SheltonEtAl2015}, instead of centering the data at the linear combination of $\sum_h s_h\vec{W}_h=W\vec{s}$.
\end{description}

In the above models, inference with the truncated posterior of Equation~\eqref{eq:sel-post} using manually-derived selection functions has yielded results as good as or more robust than exact inference (converging less frequently to local optima than exact inference; see earlier references for details).
For models \textbf{A} and \textbf{C}, the selection function was the cosine similarity between the weights $\vec{W}_h$ (e.g. dictionary elements, components, etc.) associated with each latent variable $s_h$ and 
each data point $\vec{y}^{(n)}$: 
$  \mathcal{S}_h(\vec{y}^{(n)}) = (\vec{W}_{h}^{\mathrm{T}}\,/\,||\vec{W}_{h}||)\,\vec{y}^{(n)}$.
For model \textbf{B}, the  selection function was the data likelihood given a singleton state:
 $\mathcal{S}_h(\vec{y}^{(n)}) = p(\vec{y}^{(n)} | \vec{s}=\vec{s}_h, \Theta)$, 
where $\vec{s}_h$ represents a singleton state in which only the entry $h$ is non-zero.

We generate $N=2,000$ data points consisting of $D=5\times5=25$ observed dimensions and $H=10$ latent components according to each of the models \textbf{A-C}:
$N$ images of randomly selected overlapping 'bars' with varying intensities for models \textbf{B} and \textbf{C}, and additive Gaussian noise parameterized by ground-truth $\sigma^2 = 2$ and we choose $H' = 5$, (e.g. following the spike-and-slab prior). 
On average, each data point contains $2$ bars, i.e. ground-truth is $\pi H = 2$, and we choose $H'=5$. With this choice,
we can select sufficiently many latents for virtually all data points.


For each of the models considered, we run 10 repetitions of each of the following set of experiments: (1) selection using the respective hand-crafted selection function,
(2) GP-select using a linear covariance kernel, (3) GP-select using an RBF covariance kernel, and (4) GP-select using a kernel composed by adding the following kernels: RBF, linear, bias and white noise kernels, which we will term the \emph{composition kernel}.
As hyperparameters of kernels are learned, the composition kernel (4) can adapt itself to the data and only use the kernel components required. 
See  \citet[Chapter 4, Secton 4.2.4]{RasmussenGPbook}  for a discussion on  kernel adaptation.
Kernel parameters were model-selected via maximum marginal likelihood every $10$ EM iterations.
For models \textbf{A} and \textbf{B}, inference was performed exactly using the truncated posterior~\eqref{eq:sel-post}, but as exact inference is analytically intractable in model \textbf{C}, inference was performed by drawing Gibbs samples from the truncated space \citep{SheltonEtAl2011,SheltonEtAl2012,SheltonEtAl2015}.
%
We run all models until convergence. 

\begin{figure}[h!]
\begin{center}
\includegraphics[width=.8\textwidth]{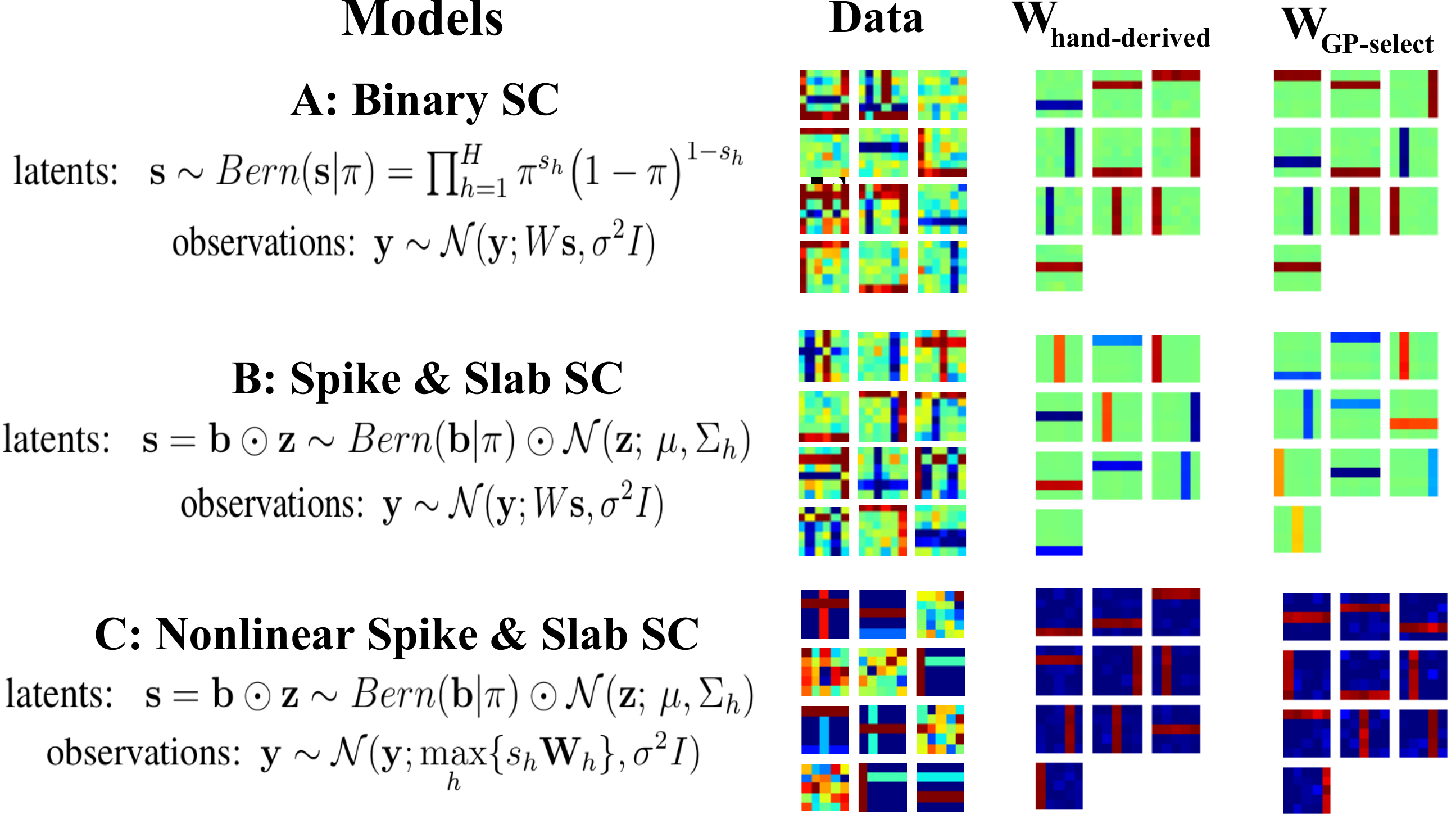}
\caption{Sparse coding models results comparing GP-select with a successful hand-derived selection function.
Results are shown on artificial ground-truth data with $H=10$ latent variables and $H'=5$ preselected variables for: \textbf{A} Binary sparse coding, \textbf{B} Spike-and-slab sparse coding, and \textbf{C} Nonlinear spike-and-slab sparse coding.
First column: Example data points $\vec{y}^{(n)}$ generated by each of the models.
Middle column: Converged dictionary elements $W$ learned by the hand-crafted selection functions.
Third column: Converged dictionary elements $W$ learned by GP-select with $H'=5$ using the kernel with best performance (matching that of inference with hand-crafted selection function).
In all cases, the model using the GP-select function converged to the ground-truth solution, just as the hand-crafted selection functions did.
}\label{fig:sparse}
\end{center}
\end{figure}

Results are shown in Figure~\ref{fig:sparse}.
In all experiments, the GP-select approach was able to infer ground-truth parameters as well as the hand-crafted function.
For models where the cosine similarity was used (in \textbf{A} and \textbf{C}), GP regression with a linear kernel quickly learned the ground-truth parameters, and hence fewer iterations of EM were necessary.
In other words, even without providing GP-select explicit weights $W$ as required for the hand-crafted function, its affinity function using GP regression~\eqref{eq:gp-loo} learned a similar enough function to quickly yield identical results.
Furthermore, in the model with a less straight-forward hand-crafted function (in the spike-and-slab model of \textbf{B}), only GP regression with an RBF kernel was able to recover ground-truth parameters.
In this case (model \textbf{B}), GP-select using an RBF kernel recovered the ground-truth 'bars' in $7$ out of $10$ repetitions, whereas the hand-crafted function recovered the bases in $8$ instances.
For the remaining models, GP-select converged to the ground-truth parameters with the same average frequency as the hand-crafted functions.


Finally, we have observed empirically that the composition kernel is flexible enough to subsume all other kernels:
the variance of the irrelevant kernels dropped to zero in simulations.
This suggests the composition kernel is a good choice  for general use. 

\subsection{Gaussian mixture model}
Next, we apply GP-select to a simple example, a Gaussian mixture model, where the flexibility of the approach can be easily and intuitively visualized.
The model of the data likelihood is
\vspace{-.2cm}
\begin{equation}\label{eq:mog}
p(\vec{y}^{(n)} | \vec{\mu_c}, \vec{\sigma_c}, \vec{\pi}) = \sum_{c=1}^{C} \mathcal{N}(\vec{y}^{(n)}; \vec{\mu_c}, \vec{\sigma_c}) \, \pi_c,
\end{equation}
where $C$ is the number of mixture components; the task is to assign each data point to its latent cluster.
%

The training data used for GP regression was $\mathcal{D} = \{ (\vec{y}^{(n)}, \langle s_h^{(n)}\rangle_{q_n}) | n = 1, \dots, N \}$, where the targets were the expected cluster responsibilities (posterior probability distribution for each cluster) for all data points, $\langle s_h\rangle_{q}$, and we use one-hot encoding for cluster identity.
With this, we apply our GP-select approach to this model, computing the selection function according to Equation~\eqref{eq:sel-func} with affinity $f$ defined by GP regression~\eqref{eq:gp-loo} 
and following the approximate EM approach as in the previous experiments.
%
In these experiments we consider two scenarios for EM learning of the data likelihood in Equation~\eqref{eq:mog}: GP-select with an RBF covariance kernel and a linear covariance kernel.  
We do not include the composition kernel suggested (based on experiments) in Section 4.1, as the goal of the current experiments is to show the  effects of using the 'wrong' kernel. These effects would further support the use of the flexible composition kernel in general, as it can subsume both kernels considered in the current experiments (RBF and linear).
%

\begin{figure*}[t]
\begin{center}
\includegraphics[width=1\textwidth]{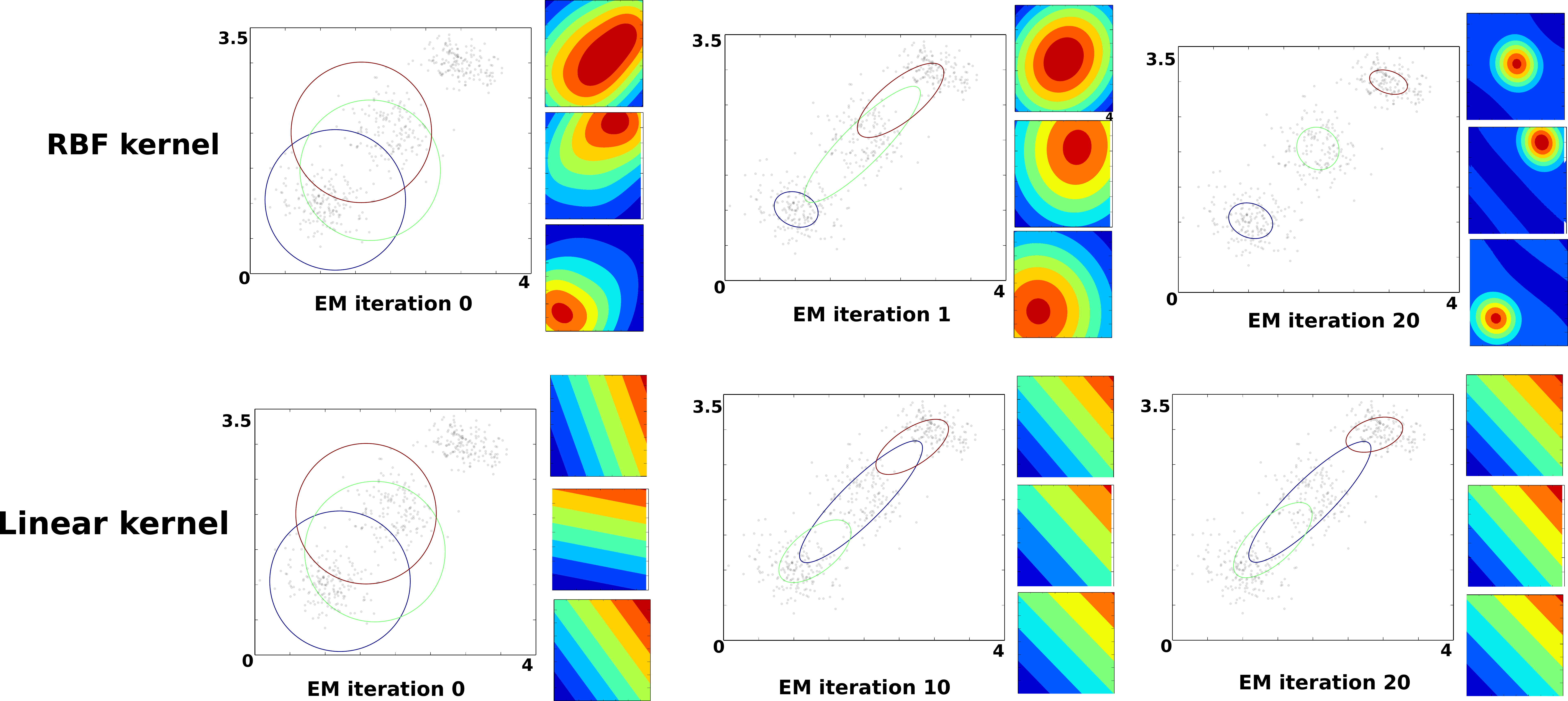}
\caption{Gaussian mixture model results using GP-select (selection of $C'=2$ in a $C=3$ class scenario) for inference.
Progress of the inference is shown using (row one) an RBF covariance kernel in the regression, and (row two) a linear covariance kernel.
For each iteration shown, we see (1) the observed data and their inferred cluster assignments and (2) the $C$ corresponding GP regression functions learned/used for GP-select in that iteration. Different iterations are pictured due to different convergence rates. As shown, inference with GP-select using a linear kernel is unable to assign the data points to the appropriate clusters, whereas GP-select with an RBF kernel succeeds.}\label{fig:mog}
\end{center}
\end{figure*}

To easily visualize the output, we generate $2$-dimensional observed data ($\vec{y}^{(n)} \in \mathrm{R}^{D=2} $) from $C=3$ clusters -- first with randomly assigned cluster means, and second such that the means of the clusters lie roughly on a line.
In the GP-select experiments, we select $C' = 2$ clusters from the full set,
and run $40$ EM iterations for both kernel choices (linear and RBF).
Note that for mixture models, the notation of $C'$ selected clusters of the $C$ set is analogous to the $H'$ selected latent variables from the $H$ full set, as described in the non-mixture model setting, and the GP-select algorithm proceeds unchanged.
We randomly initialize the variance of the clusters $\vec{\sigma_c}$ and initialize the cluster means $\vec{\mu_c}$ at randomly selected data points.
Results are shown in Figure~\ref{fig:mog}.

With cluster parameters initialized randomly on these data, the linear GP regression prediction cannot correctly assign the data to their clusters (as seen in Figure~\ref{fig:mog}\textbf{B}), but the nonlinear approach successfully and easily finds the ground-truth clusters (Figure~\ref{fig:mog}\textbf{A}).
Furthermore, even when both approaches were initialized in the optimal solution, the cluster assignments from GP-select with a linear kernel quickly wandered away from the optimal solution and were identical to random initialization, converging to the same result shown in iteration 20 of Figure~\ref{fig:mog}\textbf{B}).
The RBF kernel cluster assignments remained at the optimal solution even with number of selected clusters set to $C'=1$.


These experiments demonstrate that the selection function needs to be flexible even
for very simple models, and that nonlinear selection functions are an essential tool
even in such apparently straightforward cases.


%


\subsection{Translation Invariant Occlusive models}
\label{invec}


Now that we have verified that GP-select can be applied to various generative graphical models and converge to ground-truth parameters, we consider a more challenging model that addresses a problem in computer vision: \emph{translations of objects in a scene}.

\textbf{Model.}
Translation invariant models are particularly difficult to optimize because they must consider a massive latent variable space: evaluating multiple objects and locations in a scene leads a \textit{latent space complexity of the number of locations exponentiated by the number of objects}.
Inference in such a massive latent space heavily relies on the idea of variable selection to reduce the number of candidate objects and locations. In particular, hand-engineered selection functions that consider \emph{translational invariance} have been successfully applied to this type of model~\citep{DaiLucke2012b,DaiLucke2014,DaiEtAl2013}.
The selection function used so far for reducing latent space complexity in this model has been constructed as follows: first, the candidate locations of all the objects in the model are predicted, and then a subset of candidate objects that might appear in the image are selected according to the predicted locations.  Next, the subset of states $\Kn$ is constructed according to the combinatorics of the different locations of all the candidate objects.
The posterior distribution is then computed following Equation~\eqref{eq:sel-post}.

This selection system is very costly: the selection function has parameters which need to be hand-tuned, e.g., the number of representative features, and 
it needs to scan through the entire image, considering all possible locations, which becomes computationally demanding for large-scale experiments.
%
In translation invariant models, instead of predicting the existence of a component, the selection function has to predict all possible locations a component could be.
To maximally exploit the capabilities of GP-selection function, we directly use the GP regression model to \textit{predict the possible locations} of a component \textit{without introducing any knowledge of translation invariance} into the selection function. In this work, a GP regression model is fitted from the input image to marginal posterior probabilities of individual components appearing at all possible locations. Therefore, the input to the GP-selection function is the image to be inferred and the output is a score for each possible location of each component in the model.
For example, when learning $10$ components in a $D=30\times 30$ pixel image patch, the output dimensionality of GP-select is $9000$.
This task is computationally feasible, since GP models scale linearly with output dimensionality.
The inference of components' locations with GP-select is significantly faster than the selection function in the original work, as it avoids explicitly scanning through the image.

Although there are additional computations necessary for an automatic selection function like GP-select, for instance due to the adjustment of its parameters,  
there are many options to reduce computational costs. 
First, we may approximate the full $N \times N$ Gram matrix by an incomplete Cholesky approximation~\citep{FinSch01} 
resulting in a cost of $O(N\times Q)$, where $Q << N$ is the rank of the Cholesky approximation.
%
Second, we may reduce the frequency of kernel hyperparameter updates to only every $T^*$ EM iterations, where a $T^* > 1$ represents a corresponding computation reduction.
The combination of the Cholesky approximation plus infrequent updates will have the following benefits: a factor of five speedup for infrequent updates, and a factor of $(N-Q)^2$ speedup from incomplete Cholesky, where $Q$ is the rank of the Cholesky approximation and $N$ is the number of original data points. 


\textbf{COIL Dataset.}
\begin{figure*}[t!]
\centering
\includegraphics[width=.5\textwidth]{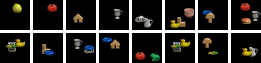}
\caption{COIL Dataset \citep{coil100}: 
A handful of data points used in experiments with the Translation Invariant Occlusive (InvECA) model, showing the occluding objects to be learned.
}
\label{fig:inveca-data}
\end{figure*}
We apply our GP-selection function to the \emph{Invariant Exclusive Component Analysis (InvECA) model}~\citep{DaiLucke2012b,DaiEtAl2013}.
For our experiments, we consider an image dataset used in previous work: data were generated using objects from the COIL-100 image dataset \citep{coil100}, taking $16$ different objects, downscaled to $D=10 \times 10$ pixels and segmented out from the black background.
A given image was generated by randomly selecting a subset of the $16$ objects, where each object has a probability of $0.2$ of appearing.
The appearing objects were placed at random positions on a $30 \times 30$ black image.
When the objects overlap, they occlude each other with a different random depth order for each image.
In total, $N=2000$ images were generated in the dataset (examples shown in Figure~ \ref{fig:inveca-data}).

The task of the InvECA model is to discover the visual components (i.e. the images of $16$ objects) from the image set without any label information. 
We compare the visual components learned by using \emph{four different selection functions in the InvECA model}: the hand-crafted selection function used in the original work by ~\citet{DaiLucke2012b}, GP-select updated every iteration, GP-select updated every $T*=5$ iterations, and GP-select with incomplete Cholesky decomposition updated every iteration, or $T*=1$ (in this manner we isolate the improvements due to Cholesky from those due to infrequent updates). 
In these experiments, the parameters of GP-select are optimized at the end of each $T*$ EM iteration(s), using a maximum of $20$ gradient updates.
The number of objects to be learned is $H=20$ and the algorithm pre-selects $H'=5$ objects for each data point.
The kernel used was the composition kernel, as suggested in Section 4.1, although after fitting the hyperparameters only the RBF kernel remained with large variance (i.e. a linear kernel alone would not have produced good variable selection, thus the flexible composition kernel was further shown to be a good choice).

\textbf{Results.}
%
\begin{figure*}[t!]
\centering
\includegraphics[width=\textwidth]{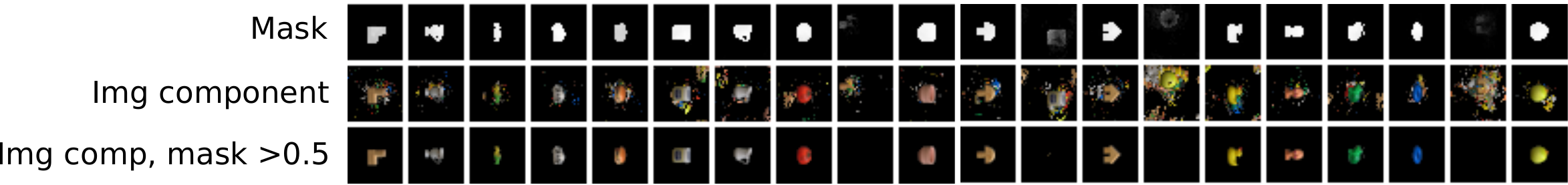}
\caption{Image components and their masks learned by GP-select with the Translation Invariant model. GP-select learned all objects in the dataset.
The first row shows the mask of each component, 
the second row shows the learned image components, 
and the third row shows only the area of the learned components that had a mask $>0.5$.
}
\label{fig:inveca-params}
\end{figure*}
All four versions of the InvECA model using each of the selection functions considered successfully recover each individual objects in our modified COIL image set. 
The learned object representations with GP-select are shown in Figure~\ref{fig:inveca-params}. 
Four additional components developed into representations, however these all had very low mask values, allowing them to easily be distinguished from other true components.

\begin{figure}[ht!]
\centering
\includegraphics[width=\textwidth]{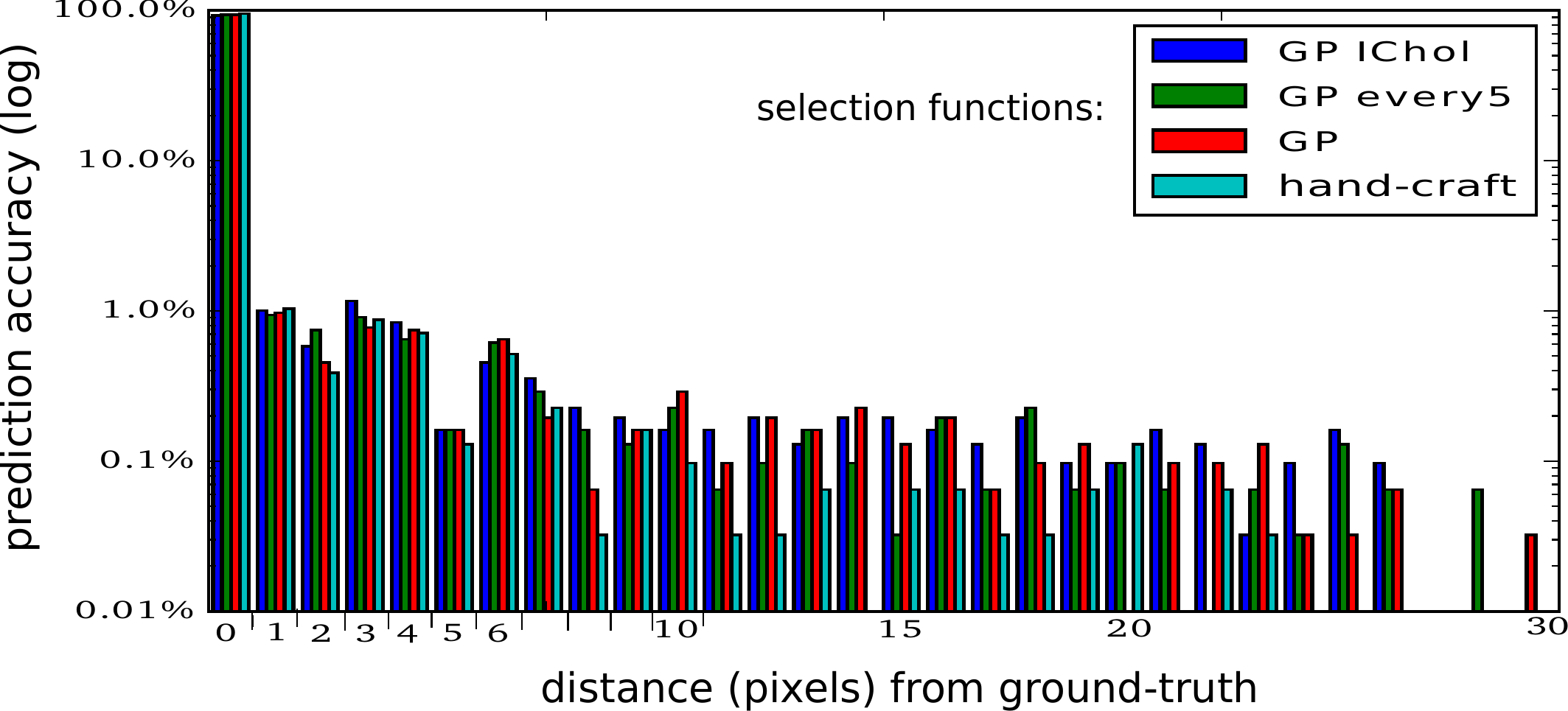}
\caption{Prediction accuracy of the four selection functions in the InvECA model.
Functions depicted in the figures: GP-select with no modifications (GP, red), the incomplete Cholesky decomposition (GP IChol, blue), with updated kernel hyperparameters every 5 EM iterations (GP every5, green), and with hand-crafted selection (hand-craft, cyan).
Shown: the log-scale histogram of the prediction accuracy for the four selection functions, measured by the distance each function's predicted object location was to the ground-truth object location. 
All bars of the selection functions show very similar accuracy for the various distances. 
}
\label{fig:inveca1}
\end{figure}
\begin{figure}[ht!]
\centering
\includegraphics[width=.6\textwidth]{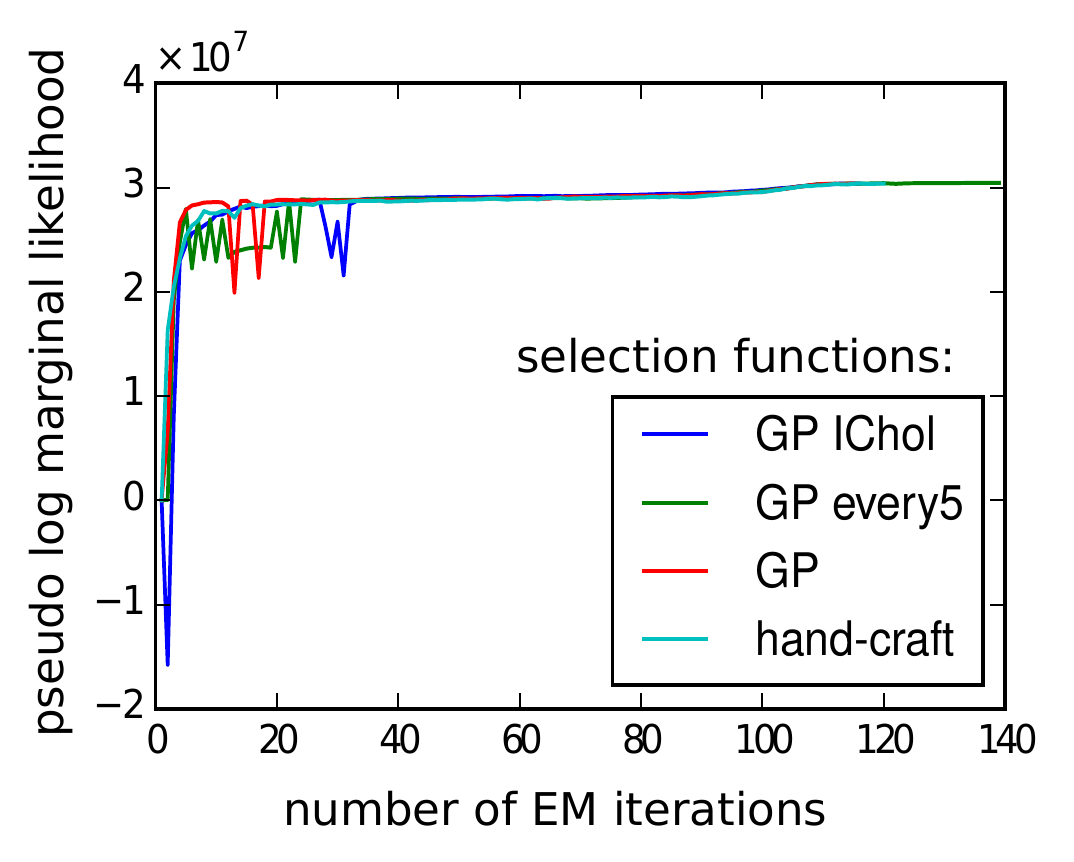}
\caption{Baseline comparison of the four selection functions in the InvECA model.
Functions depicted in the figures are identical to those in Figure~\ref{fig:inveca1}.
Shown: the convergence of the pseudo log marginal likelihood [of the model parameters learned at each EM iteration] for the four selection functions over all EM iterations. After about 40 EM iterations, all selection function versions of the algorithm converge to the same likelihood solution.  
Simultaneously, the GP-select approaches exhibit no loss of accuracy compared to the hand-crafted function,  and 
'GP IChol' represents a factor of $100$ speedup vs. 'GP', and 'GP every5' represents a factor of $5$ speedup.
}
\label{fig:inveca2}
\end{figure}
Next, we compare the accuracy of the four selection functions.
For this, we collected the object locations (pixels) indicated by each selection function after all EM iterations, 
applied the selection functions (for the GP selection functions, this was using the final function learned after all EM iterations) to the entire image dataset again, 
then compared these results with the ground-truth location of all of the objects in the dataset.
%
%
The accuracy of the predicted locations was then computed by comparing the distance of all ground-truth object location to the location of the top candidate locations from each selection function. 
 See Figure~\ref{fig:inveca1} for a histogram of these distances and the corresponding accuracy for all selection functions. Note that the percentages in the histogram are plotted in log scale.
Also, as a baseline verification, 
%
we computed and compared the pseudo log likelihood \citep{DaiEtAl2013} of the original selection function to the three GP-select based ones.
The pseudo log likelihood for all selection functions is shown in Figure~\ref{fig:inveca2}.
Figures~\ref{fig:inveca1}-\ref{fig:inveca2} show that all four selection functions can very accurately predict the locations of all the objects in the dataset -- 
the GP-select selection functions yields no loss in inference performance in comparison to the original hand-engineered selection function. 
Even those using speed-considerate approximations (incomplete Cholesky decomposition of the kernel matrix (GP IChol) and updating kernel hyperparameters only every 5 EM iterations (GP every5)) have indistinguishable prediction accuracy on the task.


An analysis of the benefits indicate that, as GP-select avoids explicitly scanning through the image, the time to infer the location of an object is significantly reduced compared to the hand-crafted function. GP-select requires $22.1$ seconds on a single CPU core to infer the locations of objects across the whole image set, while the hand-crafted function requires $1830.9$ seconds. In the original work, the selection function was implemented with GPU acceleration and parallelization. 
Although we must compute the kernel hyperparameters for GP-select, 
it is important to note that the hyperparameters need not be fit perfectly each iteration -- for the purposes of our approach, a decent approximation suffices for excellent variable selection. 
 In this experiment, updating the parameters of GP-select with $10$ gradient steps took about $390$ seconds for the full-rank kernel matrix. 
When we compute the incomplete Cholesky decomposition while inverting the covariance matrix, compute time was reduced to $194$ seconds (corresponding to the $(N-Q)^2$ speedup, where $Q$ is the rank of the Cholesky approximation), with minimal loss in accuracy.
Furthermore, when updating the GP-select hyperparameters only every 5 iterations, average compute time was reduced by another one fifth, again without loss in accuracy.


\section{Discussion}
\label{disc}
%
We have proposed a means of achieving fast EM inference in Bayesian generative models, by
learning an approximate selection function to determine relevant latent variables
for each observed variable. The process of learning the relevance functions
is interleaved with the EM steps, and these functions
are used in obtaining an approximate posterior distribution in the subsequent EM iteration.
The functions themselves are learned via Gaussian process regression,
and do not require domain-specific engineering, unlike previous selection functions.
In experiments on mixtures and sparse coding models with interpretable output,
the learned selection functions behaved in accordance with our expectations for the posterior
distribution over the latents.  

The significant benefit we show empirically is that by learning the selection function in a general and flexible nonparametric way, we can avoid using expensive hand-engineered selection functions.
Cost reduction is both in terms of required expertise in the problem domain, and computation time in identifying the relevant latent variables.
Inference using our approach required 22.1 seconds on a single CPU core, versus  1830.9 seconds with the original hand-crafted function 
for the complex hierarchical model of \citep{DaiEtAl2013}.

A major area where further performance gains might be expected is in
improving computational performance, since we expect the greatest
advantages of GP-select to occur for complex models at large scale. For instance,
 kernel ridge regression may be parallelized \citep{zhang14divide},
or the problem may be solved in the primal via random Fourier features \citep{LeSarSmo13}.
Furthermore, there are many recent developments regarding the scaling up of GP inference to large-scale problems, e.g., sparse GP approximation
~\citep{sparseGP}, stochastic variational inference \citep{HensmanEtAl2013,Hensman2012}, using parallelization techniques and GPU acceleration \citep{DaiEtAl2014}, or in combination with stochastic gradient descent~\citep{Bottou08thetradeoffs}. 
For instance, for very large datasets where the main model is typically trained with mini-batch learning, stochastic variational inference can be used for GP fitting as in \citep{HensmanEtAl2013} and the kernel parameters can be efficiently updated each (or only every $T*$ few) iteration with respect to a mini-batch.

\subsection*{Acknowledgments}
We acknowledge funding by the German Research Foundation (DFG) under grant LU 1196/4-2 (JS), by the Cluster of Excellence EXC 1077/1 "Hearing4all" (JL), and by the RADIANT and WYSIWYD (EU FP7-ICT) projects (ZD).


\bibliographystyle{apa}
\bibliography{gp}

\end{document}